%% file: main_v4.tex
\documentclass[letterpaper, 10 pt, conference, twoside]{ieeeconf}  % Comment this line out if you need a4paper

\IEEEoverridecommandlockouts                              % This command is only needed if you want to use the \thanks command
\input{utils/preamble}
\input{utils/acronyms}
\input{\MainPath/utils/shortcuts.tex}

\title{\LARGE \bf
\varTitle: Local Navigation \revised{Beyond} Observation via Conditional Flow Matching in the Latent Space
}

\author{Jiwon Park$^{1\dagger}$, Dongkyu Lee$^{2,3\dagger}$, 
  I Made Aswin Nahrendra$^{2,4}$, Jaeyoung Lim$^{5}$ and Hyun Myung$^{2*}$%
  \thanks{$^\dagger$These authors contributed equally.}%
  \thanks{$^*$Corresponding author: Hyun Myung.}%
  \thanks{$^1$Robotics Program, KAIST (Korea Advanced Institute of Science and Technology), Daejeon 34141, South Korea.
    {\tt\scriptsize ziwon@kaist.ac.kr}}%
  \thanks{$^2$School of Electrical Engineering, KAIST (Korea Advanced Institute of Science and Technology), Daejeon 34141, South Korea.
    {\tt\scriptsize \{dklee, anahrendra, hmyung\}@kaist.ac.kr}}%
  \thanks{$^3$URobotics, Seoul, South Korea. {\tt\scriptsize dklee@urobotics.ai}}%
  \thanks{$^4$KRAFTON, Seoul, South Korea. {\tt\scriptsize anahrendra@krafton.com}}%
  \thanks{$^5$Dept.\ of Electrical Engineering and Computer Science, UC Berkeley, USA. 
    {\tt\scriptsize jaeyounglim@berkeley.edu}}%
}
\begin{document}
\maketitle

\IEEEpeerreviewmaketitle

\input{\MainPath/0_Abstract/abstract}

\input{\MainPath/1_Introduction/introduction}
\input{\MainPath/2_Related_Work/related_work}
\input{\MainPath/3_Methodology/methodology}
\input{\MainPath/4_Experimental_Setup/setup}

\input{\MainPath/5_Results_and_Analysis/results}
\input{\MainPath/6_Conclusion/conclusion}

% \bibliographystyle{IEEEtran}
% \bibliography{./bibliography/reference,./utils/IEEEabrv}
\bibliographystyle{./utils/IEEEtran}
\bibliography{./bibliography/abrv,./bibliography/reference}

\end{document}

%% file: utils/preamble.tex
\makeatletter
\let\NAT@parse\undefined
\makeatother
\usepackage[T1]{fontenc}
\usepackage{cite}
\usepackage{amssymb,amsfonts}
\usepackage{hyperref}
\usepackage{algorithm}
\usepackage{algorithmicx}
\usepackage{algpseudocode}
\usepackage{graphicx}
\usepackage{textcomp}
\usepackage{mathtools}
\usepackage{changes}
\usepackage[font=footnotesize,labelformat=simple]{subcaption}
\usepackage{xcolor}
\usepackage{color, soul}
\usepackage{transparent}
\usepackage{amsmath}
\usepackage{booktabs}
\usepackage{multirow}
\usepackage{xstring}
\usepackage{hhline}
\usepackage{kotex}
\usepackage{siunitx}
\usepackage{comment}
\usepackage{physics}
\usepackage{tikz}
\usepackage{cleveref}
\usepackage{gensymb}
\usepackage{threeparttable}
\usepackage{caption}
\usepackage{bm}
\usepackage{./utils/macros}
\usepackage{adjustbox}
\usepackage{acro}
\usepackage{siunitx}    % S[table-format=4.2]
\usepackage[font=small]{caption}

\newcommand{\etal}{\textit{et al.}\ }

%% file: utils/acronyms.tex
\DeclareAcronym{CFM}{
  short = CFM,
  long  = conditional flow matching,
  short-indefinite = a,
  long-indefinite = a
}

\DeclareAcronym{RL}{
  short = RL,
  long  = reinforcement learning,
  short-indefinite = an,
  long-indefinite = a
}

\DeclareAcronym{DRL}{
  short = DRL,
  long  = deep reinforcement learning,
  short-indefinite = a,
  long-indefinite = a
}

\DeclareAcronym{FM}{
  short = FM,
  long  = flow matching,
  short-indefinite = an,
  long-indefinite = a
}

\DeclareAcronym{ODE}{
  short = ODE,
  long = ordinary differential equations,
  short-indefinite = an,
  long-indefinite = an
}

\DeclareAcronym{CNF}{
   short = CNF,
   long = continuous normalizing flows,
   short-indefinite = a,
   long-indefinite = a
}

\DeclareAcronym{PPO}{
  short = PPO,
  long  = proximal policy optimization,
  short-indefinite = a,
  long-indefinite = a
}

\DeclareAcronym{VAE}{
  short = VAE,
  long  = variational autoencoder,
  short-indefinite = a,
  long-indefinite = a
}

\DeclareAcronym{CENet}{
  short = CENet,
  long  = context-aided estimator network,
  short-indefinite = a,
  long-indefinite = a
}

\DeclareAcronym{IMU}{
  short = IMU,
  long  = inertial measurement unit,
  short-indefinite = an,
  long-indefinite = an
}

\DeclareAcronym{ID}{
  short = ID,
  long  = in-distribution,
  short-indefinite = an,
  long-indefinite = an
}

\DeclareAcronym{OOD}{
  short = OOD,
  long  = out-of-distribution,
  short-indefinite = an,
  long-indefinite = an
}

\DeclareAcronym{MLP}{
  short = MLP,
  long  = multi-layer perceptron,
  short-indefinite = an,
  long-indefinite = a
}

%% file: v4/utils/shortcuts.tex
\newcommand{\varTitle}{DreamFlow}

\newcommand{\timeStep}{t}
\newcommand{\navPolicy}{\mathbf{\pi}_{\text{nav}}}
\newcommand{\varAction}{\mathbf{a}_\timeStep}
\newcommand{\varObs}{\mathbf{o}^\mathrm{p}_\timeStep}
\newcommand{\varHeightObs}{\mathbf{o}^\mathrm{e}_\timeStep}
\newcommand{\varState}{\mathbf{s}^\mathrm{p}_\timeStep}
\newcommand{\varHeightState}{\mathbf{o}^\mathrm{E}_\timeStep}

\newcommand{\varLatent}{\mathbf{z}}
\newcommand{\varLatentLocal}{\varLatent^\text{e}_\timeStep}
\newcommand{\varLatentExtended}{\varLatent^\text{E}_\timeStep}
\newcommand{\varLatentPredicted}{\mathbf{\hat{z}}^\mathrm{e}_\timeStep}

\newcommand{\FMstep}{\tau}
\newcommand{\condition}{\mathbf{c}}
\newcommand{\conditionT}{\condition_\timeStep}
\newcommand{\velocityField}{v_{\theta}(\FMstep, \varLatent)}
\newcommand{\condVelocityField}{v_{\theta}(\FMstep, \conditionT, \varLatent)}
\newcommand{\condVelocityFieldFM}{v_{\theta}(\FMstep, \conditionT, \varLatent_\FMstep)}
\newcommand{\KcondVelocityFieldFM}{v_{\theta}(\FMstep_k, \conditionT, \varLatent_{\FMstep_k})}

\newcommand{\initDist}{p_0}
\newcommand{\targetDist}{q(\varLatent)}
\newcommand{\initDistCond}{p_0(\conditionT)}
\newcommand{\targetDistCond}{q(\varLatent|\conditionT)}

\newcommand{\targetVelocityField}{\mathbf{u}_\FMstep(\varLatent|\conditionT)}

\newcommand{\locEncoder}{\mathcal{H}^\mathrm{e}}
\newcommand{\extEncoder}{\mathcal{H}^\mathrm{E}}
\newcommand{\trainingData}{\mathcal{D}}

%% file: v4/0_Abstract/abstract.tex
\begin{abstract}
% Autonomous navigation in cluttered environments is challenging due to dense obstacles and frequent local minima. 
Local navigation in cluttered environments often suffers from dense obstacles and frequent local minima.
Conventional local planners rely on heuristics and are prone to failure, while deep reinforcement learning (DRL)-based approaches provide adaptability but are constrained by limited onboard sensing. 
These limitations lead to navigation failures because the robot cannot \revised{perceive} structures outside its field of view.
In this paper, we propose \varTitle, a DRL-based local navigation framework that extends the robot’s perceptual horizon through conditional flow matching (CFM).
The proposed CFM-based prediction module learns probabilistic mapping between local height map latent representation and broader spatial representation conditioned on navigation context.
This enables the navigation policy to predict unobserved environmental features and proactively avoid potential local minima.
Experimental results demonstrate that \varTitle \space outperforms existing methods in terms of latent prediction \revised{accuracy} and navigation performance in simulation. 
The proposed method \revised{was} further validated in cluttered real-world environments with a quadrupedal robot.
The project page is available at \url{https://dreamflow-icra.github.io}.
\end{abstract}

%% file: v4/1_Introduction/introduction.tex
\section{Introduction}
\label{sec:introduction}

Safe navigation in cluttered environments is a key requirement for autonomous mobile systems, including search and rescue~\cite{niroui2019deep}, exploration~\cite{kan2020online}, and environmental monitoring~\cite{kim2022autonomous}.
Such environments contain dense obstacles and severe visual occlusions, which make reaching target locations highly challenging and call for structured navigation strategies. 
To address this, hierarchical navigation pipelines ~\cite{jian2021global, chu2012local} have been widely adopted, decomposing navigation into perception, planning, and control.
Among them, the role of the planning module is to generate feasible commands that guide the robot toward its goal, based on the environmental representation constructed by the perception module.
% The perception module processes sensor data to construct a  representation of the surrounding environment, often incorporating mapping capabilities~\cite{chen2025learning, fankhauser2018probabilistic}. 
% The planning module then generates feasible commands to guide the robot toward its goal, based on the environmental representation provided by perception.

\input{\MainPath/1_Introduction/fig_01.tex}

The planning module is typically divided into global planners~\cite{lee2025trg,yang2022far,liu2023hybrid}, which compute long-horizon paths for high-level guidance, and local planners~\cite{nakhleh2023sacplanner,lu2023lpnet, rudin2022advanced}, which operate over shorter horizons to react to immediate disturbances and obstacles. 
However, the absence of reliable maps in real-world scenarios makes global planning impractical, especially in previously unseen or rapidly changing environments.
% ~\cite{yan2022mapless, jang2021hindsight}.
In such cases, navigation often relies on \revised{the} local navigation, which emphasizes reactive obstacle avoidance and short-horizon goal reaching under limited visibility.

A key challenge in local navigation is \revised{the} \textit{local minima} problem, a situation where the robot becomes trapped because no available action leads closer to the goal.
% ~\cite{kim2025escaping, meijer2025pushing}.
% For example, when the robot faces a U-shaped obstacle or a dead-end corridor, moving forward risks collision while turning back seems to move away from the target.
Typical cases include U-shaped obstacles or dead-end corridors, where progressing forward risks \revised{a} collision while retreating increases the distance to the target. 
Such situations arise because the \revised{local} planner considers only the robot’s immediate perceptual horizon, prioritizing short-term safety over long-term goal progress.
As a result, the robot may misjudge the goal as unreachable even though a feasible path exists beyond its field of view.
Conventional local planners rely on simple heuristics~\cite{conner2003composition, klanvcar2021combined, fuke2025towards} to escape such traps, but \revised{frequently} fail in cluttered or highly occluded environments.

Learning-based methods~\cite{kim2025escaping, meijer2025pushing, guldenring2020learning, patel2021dwarl, liu2020robot} have recently emerged as a promising alternative \revised{to} local navigation, with \ac{DRL} receiving particular attention for its ability to train policies directly through interaction with the environment while incorporating robot \revised{kino-dynamics} and feedback signals.
% Given the limitations of heuristic approaches, learning-based methods have emerged as a promising alternative for local navigation.
% Especially, \ac{DRL} has attracted particular attention because it allows policies to be trained directly through interaction with the environment, while incorporating robot kinodynamics and feedback signals~\cite{guldenring2020learning, patel2021dwarl, liu2020robot}.
These approaches reduce reliance on manually designed heuristics and \revised{enable} more adaptive behaviors in cluttered environments.
However, \revised{a} \ac{DRL} policy still relies on a partially observable map acquired through \revised{an} onboard perception system.
This limited perception often prevents the robot from reasoning beyond its immediate view, leaving it vulnerable to local minima in complex environments.
Although some recent works \revised{have improved} navigation performance by incorporating robot dynamics over time~\cite{hoeller2021learning} \revised{and} by reconstructing more complete environment states from partial observations~\cite{zhang2024resilient, nahrendra2026dreamwaq++}, they still rely on planning within limited local areas.

% To mitigate this issue, recent works incorporate temporal information, either by learning latent representations that capture dynamics over time~\cite{hoeller2021learning}, or by reconstructing more complete environment states from incomplete observations~\cite{zhang2024resilient}.
% Although these learning strategies improve local navigation performance, they only consider limited local areas when planning.

In this context, we propose \varTitle, a \ac{DRL}-based local navigation framework that extends the implicit perceptual range by leveraging the emerging generative model\revised{,} \ac{CFM}~\cite{lipman2023flow, liu2023flow, dao2023lfm, tong2024improving}, as illustrated in~\fig{01_main}.
% and thereby effectively avoid local minima. 
% CFM captures the inherent uncertainty of unobserved regions by learning probabilistic mappings between local and global spatial representations~\cite{tong2024improving, dao2023lfm}.
The proposed \ac{CFM}-based prediction module captures the inherent uncertainty of unobserved regions by learning probabilistic mappings between local and global representations, conditioned on navigation context.
Specifically, the \revised{network learns} optimal transport flows \revised{that} map latent representations from local height maps to those of wider regions.
Then, we \revised{integrate} the proposed \ac{CFM} prediction module into the \ac{DRL}-based navigation pipeline adapted from~\cite{zhang2024resilient}.
Therefore, the policy reasons about unobserved environmental features and proactively avoids potential local minima.
% By learning optimal transport flows to trans form latents from local heightmaps into latents from wider regions, CFM enables our RL-based planner to reason about unobserved environmental structure and proactively avoid potential traps during navigation.
% Furthermore, we condition the CFM on robot state variables alongside local latents, which enables more precise vector field learning and significantly enhances extended latent prediction.
% Furthermore, by conditioning CFM on the robot state, it learns more precise vector fields and improves extended latent prediction.
% We evaluate our approach in both simulation and real hardware experiments, demonstrating that navigation policies utilizing extended latent representations achieve up to XX\% improvement in success rate (SR) and YY\% higher success weighted by path length(SPL)~\cite{anderson2018evaluation}, particularly in complex environments prone to local minima.
% 구체적인 수치
% We evaluate our approach in both simulation and real hardware experiments, demonstrating that navigation policies utilizing extended latent representations consistently outperform baselines, achieving 6–54 percentage points higher SR and 0.08–0.37 absolute gains in SPL)~\cite{anderson2018evaluation} across different environments, particularly in scenarios prone to local minima.
% 추상적 수치

We evaluate\revised{d} our proposed method in both simulation and real-world experiments.
The results demonstrate that local navigation policies leveraging prediction of extended latent representations achieve improved navigation performance and robustness, particularly in complex environments prone to local minima.
In summary, the main contributions of this paper are as follows:
\begin{itemize}
    \item 
        We introduce \varTitle, a \ac{DRL}-based local navigation framework that addresses the local minima problem arising from the limited range of onboard sensors.
    \item 
        We propose a latent prediction module leveraging \ac{CFM} to predict extended environmental representations from local height maps under contextual conditions.
    \item 
        We demonstrate the effectiveness of \varTitle, which outperforms existing methods in simulation and is further validated in real-world environments with cluttered obstacles and local minima.

    % \item We propose a novel RL-based local navigation framework that leverages CFM to predict extended environmental representations. This allows the policy to probabilistically reason about unobserved regions and effectively avoid local minima.
    % \item We demonstrate that CFM is highly effective for this task compared to alternative learning architectures, and that conditioning CFM on robot state variables further improves the accuracy of extended latent prediction.
    % \item We validate our framework through simulation and real-world experiments, showing consistent performance gains in cluttered environments with obstacles and local minima.
\end{itemize}

%% file: v4/1_Introduction/fig_01.tex
\begin{figure}[!t]
	\captionsetup{font=footnotesize}
        % \vspace{-0.2cm}
	\centering 
        % \hfill
	\includegraphics[width=\columnwidth]{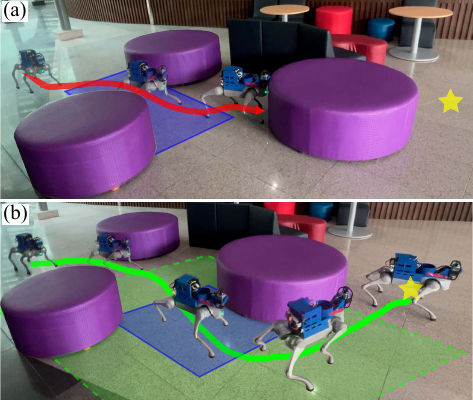} 
        \caption{
            % (a) Local navigation with only a local sensing range often fails to avoid nearby obstacles due to myopic perception, with the collision point is highlighted red.
            % (b) \varTitle \space enables the robot to find optimal actions within the limited onboard sensing range through \ac{CFM}-based prediction of extended latent representations.
            % The red and green lines indicate the robot's executed trajectories, respectively.
            \revised{An illustration of local navigation to the goal point (yellow star) in a cluttered environment with obstacles.}
            (a) Local navigation with only a limited onboard sensing range (blue box\revised{ed region}) may \revised{lead to local minima when obstacles lie beyond the sensor range.}
            % fail to avoid obstacles beyond the sensor range.
            (b) \varTitle \space enables the robot to find optimal actions within the observed sensing range (blue box\revised{ed region}) by leveraging \ac{CFM}-based prediction of extended latent representations (green box\revised{ed region}).
            The red and green lines indicate the executed trajectories, respectively.
        }
	\label{fig:01_main}
        \vspace{-0.6cm}
\end{figure}

%% file: v4/2_Related_Work/related_work.tex
\section{Related Works}\label{section:related_work}

\input{\MainPath/2_Related_Work/a.local_navigation_2}
\input{\MainPath/2_Related_Work/b.flow_matching}

%% file: v4/2_Related_Work/a.local_navigation_2.tex
\subsection{Local Navigation}

The aim of local navigation is to guide robots safely around obstacles \revised{and toward} designated goals\revised{, using} only local perception. 
Classical reactive methods~\cite{fox2002dynamic, conner2003composition, klanvcar2021combined, fuke2025towards} provide computationally efficient solutions but are prone to myopic decisions, \revised{often leading} to suboptimal paths in cluttered environments.

Recently, learning-based approaches have been proposed to address the local minima problem.
Some research~\cite{meijer2025pushing, jang2021hindsight} leverages temporal information from past interactions, enabling robots to recall failed attempts and avoid repetitive behaviors, while other works~\cite{fuke2025towards, debnath2025hybrid} integrate global planning guidance into local navigation through cost map potentials or intermediate waypoints.
However, temporal methods struggle \revised{against} novel obstacle configurations unseen during training, and hybrid approaches rely on simplified environment representations that reduce \revised{the} local planner's ability to cope with obstacles beyond the sensor range.
In this context, our approach focuses on enabling the local planner to \revised{directly predict environmental structure beyond the immediate sensor range, proactively avoiding potential traps without requiring global maps or temporal memory.}

%% file: v4/2_Related_Work/b.flow_matching.tex
\subsection{Flow Matching in Robotics}

\Ac{FM} is a simulation-free generative modeling framework that learns \ac{CNF} \revised{via} neural \ac{ODE} to transport probability distributions from source to target spaces~\cite{lipman2023flow, albergo2023building}.
Unlike diffusion models~\cite{ho2020denoising}, which rely on iterative denoising processes that \revised{are} computationally expensive during both training and sampling, \ac{FM} offers a more direct approach by learning deterministic velocity fields \revised{to} guide the transformation between distributions.
% \ac{CFM}~\cite{tong2024improving} extends \ac{FM} by conditioning on source-target pairs, enabling more flexible transport maps that incorporate optimal transport theory principles.
\ac{CFM}~\cite{tong2024improving} extends \ac{FM} by conditioning on source–target pairs, \revised{thereby enabling} the construction of more flexible transport maps.

% Nguyen~\etal\cite{nguyen2025flowmp} utilized \ac{CFM} in behavioral cloning to learn motion fields that map sensory observations to corresponding robot actions. 
% Zhang~\etal\cite{zhang2025flowpolicy} and Zhai~\etal\cite{zhai2025vfp} applied \ac{CFM} \revised{to} manipulation tasks, with the former using consistency-based approaches for policy generation from 3D point clouds and the latter enabling multi-modal policy generation in complex manipulation scenarios.
% In \revised{the} case of autonomous mobile robot\revised{s}, Gode~\etal\cite{gode2024flownav} demonstrated how \ac{CFM} can be effectively combined with depth priors to enhance navigation efficiency. 
Recent robotics applications of \ac{CFM} have targeted trajectory-level planning and control, including behavioral cloning for motion field learning~\cite{nguyen2025flowmp}, manipulation policy generation from 3D observations~\cite{zhang2025flowpolicy, zhai2025vfp}, and depth-prior-based mobile robot navigation~\cite{gode2024flownav}.
Although these works successfully demonstrate \ac{CFM}'s effectiveness in direct policy learning by predicting future robot actions, there remains an opportunity to investigate how \ac{CFM} can enhance spatial representations for improved robot navigation, which is the main focus of this work.
Although these works successfully demonstrate \ac{CFM}’s effectiveness in direct policy learning by predicting future robot actions, there remains an opportunity to investigate how \ac{CFM} can enhance spatial representations for improved robot navigation, which is the main focus of this work.

% Although these works successfully demonstrate \ac{CFM}'s effectiveness in direct policy learning by predicting future robot actions, there remains the opportunity to explore how \ac{CFM} can enhance spatial representations for improved robotic perception and planning capabilities.

%% file: v4/3_Methodology/methodology.tex
\section{\varTitle}

% The main objective of mapless navigation is 

% We first present the RL-based mapless navigation architecture, then detail the DreamFlow module and its integration.

% The objective is to develop a mapless navigation system that takes as input the observable local heightmap and relative target positions, and outputs the velocity of the robot to reach the target position while avoiding collisions.
% We define a navigation policy $\pi_{nav}$, which takes as input a encoded latents from observations and outputs actions $\mathbf{a}_t$ at each time step $t$.

\input{\MainPath/3_Methodology/fig_02.tex}
\input{\MainPath/3_Methodology/a.overview}

\input{\MainPath/3_Methodology/b.cfm_formulation}
\input{\MainPath/3_Methodology/c.dreamflow_training}
\input{\MainPath/3_Methodology/fig_03.tex}

%% file: v4/3_Methodology/fig_02.tex
\begin{figure*}[!t]
	\captionsetup{font=footnotesize}
	\centering 
	\includegraphics[width=1.0\textwidth]{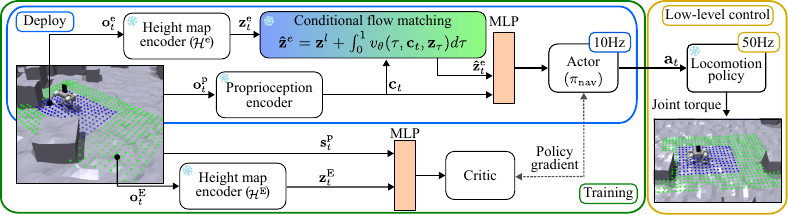}
	\caption{
        The overall framework of \varTitle.
		During deployment, \revised{the} local height map~$\varHeightObs$ (blue dots, indicating observable terrain) is encoded into \revised{the} local latent \revised{vector}~$\varLatentLocal$.
		Subsequently, the pre-trained \ac{CFM} predicts the extended latent \revised{vector}~$\varLatentPredicted$ conditioned on the context \revised{vector}~$\conditionT$ encoded from the \revised{proprioceptive} observation\revised{s}~$\varObs$.
        The predicted latent \revised{vector}~$\varLatentPredicted$ represents information from the extended height map region (green dots, indicating terrain beyond sensor range).
		The navigation policy~$\navPolicy$ takes the concatenated~$\conditionT$ and~$\varLatentPredicted$ as input to produce the high-level velocity action~$\varAction$.
		The locomotion policy then generates the low-level joint actions to control the robot.
		When training the navigation policy, the privileged states~$\varState$ and the extended latent \revised{vector}~$\varLatentExtended$ from the extended height map~$\varHeightState$ are used to train the critic network.
    }
	\label{fig:framework}
    \vspace{-0.2cm}
\end{figure*}

%% file: v4/3_Methodology/a.overview.tex
\subsection{Overall Architecture}

The objective of \varTitle \space is to enable robots to reach target positions while avoiding redundant exploration or getting stuck in local minima.
We formulate this problem as a \ac{DRL}-based navigation task, building upon the approach of Zhang~\etal\cite{zhang2024resilient}.
The overall navigation architecture, illustrated in~\fig{framework}, is designed as an asymmetric actor-critic framework~\cite{pinto2017asymmetric}.
The navigation policy~$\navPolicy$ outputs the robot's body velocity actions~$\varAction$ based on the observations, while a pretrained locomotion policy~\cite{nahrendra2023dreamwaq} is employed as a low-level controller.
Both the actor and critic networks are trained using the \ac{PPO} algorithm~\cite{schulman2017proximal}.

% The observations follow a structure similar to the baseline, consisting of the target position and heading, proprioception information such as IMU data and joint states, and exteroceptive information in the form of a height map~$\hmap{}$.
% We employ two encoders to effectively process these observations: a Variational Autoencoder (VAE) for the height map and a Context-aided Estimator Network (CENet)~\cite{nahrendra2023dreamwaq} for proprioception inputs.

A notable feature of \varTitle \space is its use of asymmetric exteroception height map observations between the actor and critic.
The actor uses a local height map~$\varHeightObs \in \mathbb{R}^{H \times W}$, which the robot can actually observe, to derive a local environmental latent representation.
In contrast, the critic utilizes an extended privileged height map~$\varHeightState \in \mathbb{R}^{H' \times W'}$, with a larger sensing range beyond the onboard sensors, to extract an extended latent representation.
Here, $H$, $W$, $H'$, and $W'$ denote the height and width of the local and extended height maps, respectively, subject to the conditions $H$ < $H'$ and $W$ < $W'$.
The proprioceptive observations~$\varObs$ and privileged states~$\varState$ contain navigation-related information such as the target position, heading command, and previous actions, along with measurements from the \ac{IMU} and joint states.

Compared \revised{with} Zhang~\etal\cite{zhang2024resilient}, the observation encoding is simplified, as our contribution mainly lies in the latent prediction module. 
Two pre-trained \acp{VAE} are utilized for height map encoder\revised{s}, $\locEncoder$ and $\extEncoder$, to extract latent representation of local and extended height map, respectively.
For proprioceptive inputs, we adopt \ac{CENet}~\cite{nahrendra2023dreamwaq}.
Subsequently, the velocity field trained via \ac{CFM}, detailed in~\subsec{cfm}, predicts the extended latent \revised{vector}~$\varLatentPredicted$ by transporting the local latent \revised{vector}~$\varLatentLocal$ toward the extended latent \revised{vector}~$\varLatentExtended$.
Finally, \revised{the} predicted latent representation \revised{vector}~$\varLatentPredicted$ and \revised{the} proprioceptive features are concatenated and fed into the actor network.

%% file: v4/3_Methodology/b.cfm_formulation.tex
\subsection{Conditional Flow Matching in Latent Space}
\label{subsec:cfm}

To overcome the limitations of partial observability in local navigation, we propose \revised{an} implicit environment prediction module using latent-to-latent \ac{CFM}~\cite{dao2023lfm}.
The proposed \ac{CFM}-based module predicts the extended latent space from the local latent space, enabling the navigation policy to make more informed decisions.

We formulate this task as the problem of learning a continuous flow that maps from \revised{the} local environmental latent \revised{vector}~$\varLatentLocal \in \mathbb{R}^{d}$ to \revised{the} extended environmental latent \revised{vector}~$\varLatentExtended \in \mathbb{R}^{d}$, where $d$ is the dimensionality of the latent space.
The $\varLatentLocal$ is an embedded vector from the limited local height map within the robot's perception range, whereas the $\varLatentExtended$ embeds \revised{a} larger spatial structure beyond \revised{the} current observ\revised{ed height map}.
Therefore, the deterministic flow, also referred to as a velocity field~$\velocityField$, is defined by the \ac{ODE} as follows:
\begin{equation}
   \label{eq:fm_ode}
   \frac{d\varLatent}{d\FMstep} = \velocityField,
\end{equation}
where $\FMstep\!\in\![0, 1]$ is the step size for interpolation between the source and target distributions, and $\varLatent$ denotes the latent representation vector.
% , and $\conditionT$ is the conditioning context, respectively.
% , and $\theta$ represents the trainable model parameters, respectively.
According to prior works on \ac{FM}~\cite{lipman2023flow, tong2024improving}, the target velocity field is intractable because it involves integrating over all pairs sampled from the Gaussian noise distribution \revised{and mapped} to the target data distribution.
However, in \varTitle, $\velocityField$ becomes learnable because we sample the initial latent~$\varLatent_0$ directly from the local latent distribution~$\initDist$, and the target latent~$\varLatent_1$ from the extended latent distribution~$\targetDist$.

Nonetheless, the prediction of \revised{the} extended latent is inherently multi-modal, as multiple extended environments \revised{might} correspond to a single local observation.
To address this, we incorporate a condition \revised{vector}~$\conditionT$ into~\eqn{fm_ode} for deterministic context-aware mapping as follows:
\begin{equation}
   \label{eq:cfm_ode}
   \frac{d\varLatent}{d\FMstep} = \condVelocityField,
\end{equation}
where $\condVelocityField$ is the conditioned velocity field\revised{,} and \revised{the} $\conditionT$ is the conditioning context that encodes the robot's state and goal information at each time step.
The solution of \eqn{cfm_ode} provides a trajectory that connects \revised{the} initial local latent representation $\varLatent_0\!=\!\varLatentLocal$ to the target extended representation $\varLatent_1\!=\!\varLatentExtended$.
In this way, the flow learns the optimal transport from an initial distribution of local latent representations~$\initDistCond$ to the target distribution of extended latent representations~$\targetDistCond$.

The conditioned velocity field~$\condVelocityField$ is optimized by minimizing the following loss function:
\begin{equation}\label{eq:flow_matching_loss}
   \mathcal{L}(\theta)=\mathbb{E}_{\FMstep, \, \varLatent_0 \sim \initDistCond, \, \varLatent_1 \sim \targetDistCond} \big[ \|\condVelocityField - \targetVelocityField\|^2 \big],
\end{equation}
where $\targetVelocityField = \varLatent_1 - \varLatent_0$ is the target velocity field under optimal transport, and $\FMstep$ is sampled from the uniform distribution $\mathcal{U}[0, 1]$.
Therefore, $\varLatent_\FMstep$ can be obtained by linear interpolation between the initial and target latent \revised{representations} as follows:
\begin{equation}
   \varLatent_\FMstep = (1-\FMstep)\varLatent_0 + \FMstep\varLatent_1.
\end{equation}

%% file: v4/3_Methodology/c.dreamflow_training.tex
\subsection{\varTitle \space Training}

The \varTitle \space training is conducted in two stages.
First, \ac{CFM} is trained to infer the prediction of \revised{the} extended latent representation from \revised{the} local observation latent \revised{representation}.
Then, the navigation policy~$\navPolicy$ is trained to output velocity commands based on the latent representation \revised{predicted} by the frozen \ac{CFM} model.

\subsubsection{CFM Training}
The objective of the \ac{CFM} training is to learn the velocity field network that transports the local latent distribution~$\initDistCond$ to the extended latent distribution~$\targetDistCond$.
We first pre-train two navigation policies using local and extended height map\revised{s}, respectively.
Then, we collect a dataset $\trainingData\!=\!\{(\varLatent^l_i, \varLatent^e_i, \condition_i)\}_{i=1}^N$ by gathering latent pairs and condition\revised{s} during multi-agent navigation episodes in \revised{the} IsaacGym~\cite{makoviychuk2021isaac} simulation.
Formally, the loss function in~\eqn{flow_matching_loss} is formulated as follows:
\begin{equation}
    \begin{aligned}
        \mathcal{L}_\text{\varTitle}(\theta) &= \mathbb{E}_{\FMstep \sim \mathcal{U}[0,1], \, (\varLatentLocal, \varLatentExtended, \conditionT) \sim \trainingData} \\
        &\quad \Big[ \| \condVelocityFieldFM - (\varLatentExtended - \varLatentLocal) \|^2 \Big],
    \end{aligned}
\end{equation}
where $\varLatent_\FMstep = (1 - \FMstep) \varLatentLocal + \FMstep \varLatentExtended$.
Therefore, $\condVelocityFieldFM$ is optimized to transport local latent \revised{vector}~$\varLatentLocal$ to extended latent \revised{vector}~$\varLatentExtended$ conditioned on~$\conditionT$, as depicted in~\fig{cfm_training}.

\subsubsection{Navigation Policy Training}
During the navigation policy training, we use the frozen \ac{CFM} to predict extended latent \revised{representations}.
The initial latent is set to the local latent representation~$\varLatentLocal$, and then the predicted extended latent representation~$\varLatentPredicted$ is generated using the \ac{CFM} velocity field~$\condVelocityFieldFM$ as follows:
\begin{equation}\label{eq:flow_matching_inference}
    \varLatentPredicted = \varLatentLocal + \int_0^1 \condVelocityFieldFM d\FMstep.
\end{equation}

In the deployment of~\eqref{eq:flow_matching_inference}, we use Euler integration with \revised{a} discrete step size~$\FMstep_k = \frac{1}{K}$:
\begin{equation}
    \varLatentPredicted \approx \varLatentLocal + \sum_{k=0}^{K-1} \FMstep_k \KcondVelocityFieldFM,
\end{equation}
where $K$ is the number of integration steps, $k \in \{0, 1, ..., K-1\}$ indexes the integration steps, and the final $\varLatentPredicted$ approximates the integral result.
The navigation policy~$\navPolicy$ is then trained using the predicted extended latent~$\varLatentPredicted$, while keeping the remaining training setup consistent with the Zhang~\etal\cite{zhang2024resilient}.
This enables the robot to exploit extended spatial representations while operating with only local sensor observations.

%% file: v4/3_Methodology/fig_03.tex
\begin{figure}[!t]
	\captionsetup{font=footnotesize}
	\centering
	\includegraphics[width=0.9\columnwidth]{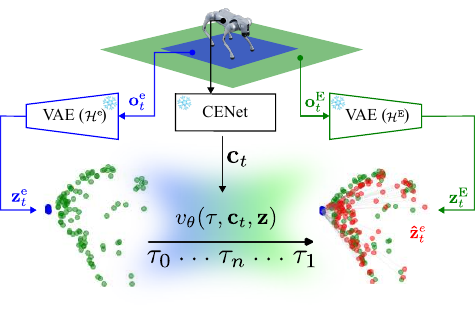}
        \caption{
        The \ac{CFM} training pipeline.
        The training dataset is collected using pre-trained height map encoders~($\locEncoder$ and $\extEncoder$) and a proprioceptive encoder (CENet).    
        \revised{A l}ocal height map~$\varHeightObs$ (blue \revised{boxed region}) \revised{is} encoded into a local latent representation~$\varLatentLocal$, and a privileged extended height map~$\varHeightState$ (green \revised{boxed region}) \revised{is} encoded into an extended latent representation~$\varLatentExtended$.
        During training, \ac{CFM} learns a velocity field~$\condVelocityField$, conditioned on the contextual vector~$\conditionT$, that transports~$\varLatentLocal$ at $\FMstep_0$ towards~$\varLatentExtended$ at $\FMstep_1$.
        Consequently, \ac{CFM} maps $\varLatentLocal$ (blue dots) to \revised{the} predicted latent \revised{vector}~$\varLatentPredicted$ (red dots), which align\revised{s} with~$\varLatentExtended$ \revised{(greed dots)} in the latent space.
        }
    \vspace{-0.2cm} %
    \label{fig:cfm_training}
\end{figure}

%% file: v4/4_Experimental_Setup/setup.tex
\section{Experiments}

We performed both simulation and real-world experiments to validate the effectiveness of the proposed method.
From a fundamental perspective, our experiments were designed to address the following questions:

\textbf{Q1: Can \ac{CFM} effectively predict \revised{an} extended latent \revised{vector} from \revised{a} partially observed local latent \revised{vector}?}
We evaluated our \ac{CFM} prediction module by measuring how accurately it predicts the target latent vector~$\varLatentExtended$ from the local latent vector~$\varLatentLocal$ compared with other model architectures~(Section~\ref{sec:results_latent_pred}).

\textbf{Q2: Do the latent predictions improve \revised{the} performance of local navigation?}
We empirically validated the assumption that the predicted latent vector~$\varLatentPredicted$ contain\revised{s} useful information about unobserved terrain through navigation experiments in simulation environments~(Section~\ref{sec:results_sim}).

\textbf{Q3: Is the proposed framework effective in real-world scenarios?}
We deployed \varTitle~in real-world environments, validating its generalization capability when exposed \revised{to} real-world uncertainties such as sensor noise and computational constraints~(Section~\ref{sec:results_real}).

\input{\MainPath/4_Experimental_Setup/a.implementation}
\input{\MainPath/4_Experimental_Setup/fig_04.tex}
\input{\MainPath/4_Experimental_Setup/b.experimental}

%% file: v4/4_Experimental_Setup/a.implementation.tex
\subsection{Implementation Details}
The simulation environment for training and evaluation was built using IsaacGym~\cite{makoviychuk2021isaac}.
During training, obstacles of varying sizes were randomly distributed on a flat terrain of size $8 \times 8 \,\mathrm{m}$.
The obstacle density was progressively increased through curriculum learning to gradually enhance the environmental complexity.
The local height map covered an area of $3 \times 2 \,\mathrm{m}$, while the privileged extended height map size was set to $6 \times 4 \,\mathrm{m}$.
Both height maps were discretized with a spatial resolution of $0.1 \,\mathrm{m}$.

The proposed \ac{CFM} model was trained on a dataset consisting of $4\mathrm{M}$ samples, with two disjoint test sets of $64\mathrm{K}$ samples each for the \ac{ID} and \ac{OOD} evaluations.
The \ac{ID} test set comprised environments similar to the training distribution, while the \ac{OOD} test set included environments with random pits, introducing terrain features not encountered during training.
The \ac{CFM} model employed a $3$-layer MLP with $256$ hidden \revised{units} and a $32$-dimensional conditioning vector.
It was trained for $1,\!000$ epochs with a batch size of $512$. During inference, the integration step $K$ for the \ac{CFM} and \revised{the} diffusion model was set to $10$.

For training the navigation policy, the height map \acp{VAE} were implemented as $3$-layer MLPs with $128$-dimensional latent spaces and pre-trained for $3,\!000$ epochs. The training took approximately $3$ and $2.5$ hours for the \ac{CFM} and navigation policy, respectively, on \revised{an} NVIDIA A5000 GPU.

%% file: v4/4_Experimental_Setup/fig_04.tex
\begin{figure}[!t]
	\captionsetup{font=footnotesize}
	\centering
	\includegraphics[width=0.9\columnwidth]{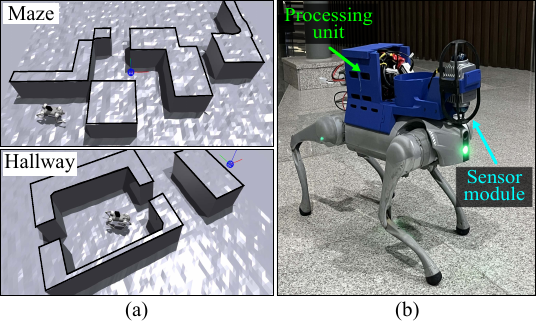}
	\caption{
            (a) The simulation environments for evaluating navigation performance. 
            The Maze environment consists of multiple corners leading to the goal, while the Hallway environment consists of narrow passages.
            (b) Our quadruped robot, used in real-world experiments, is equipped with a processing unit and a sensor module mounted on its body.
        }
	\label{fig:exp_setup}
    \vspace{-0.2cm}
\end{figure}

%% file: v4/4_Experimental_Setup/b.experimental.tex
\subsection{Experimental Setup}
\subsubsection{Latent Prediction Evaluation}
The performance of the latent prediction was evaluated \revised{using} cosine similarity between the predicted and ground-truth latent representations across \revised{the} \ac{ID} and \ac{OOD} test sets.
% The \ac{ID} test set comprised environments similar to the training distribution, while the \ac{OOD} test set included environments with random pits, introducing terrain features not encountered during training.
Additionally, the inference time of the models was evaluated on an NVIDIA 3060Ti GPU to assess computational efficiency.
We compared the latent predictions using MLP, Diffusion~\cite{ho2020denoising}, \ac{FM}, and \ac{CFM}, which was employed in \varTitle.
All models were trained on the same dataset and training settings except for \revised{differences in} the latent prediction model architecture for a fair comparison.

\subsubsection{Navigation Performance in Simulation}
For \revised{the evaluation of} navigation performance, two environments were designed: Maze and Hallway, as illustrated in \fig{exp_setup}(a).
The Maze environment is characterized by multiple corners and dead ends, requiring the robot to make strategic decisions to avoid local minima. Additionally, we set \texttt{Easy} and \texttt{Hard} scenarios according to the goal location.
In the Hallway environment, a goal is located at the end of a narrow and wall-bounded passage, which is designed to evaluate the navigation capabilities of the robot in confined spaces without collisions.
These three scenarios were performed for $1,\!000$ episodes each.

% The comparison methods for navigation performance includes:
We compared \varTitle \space against three \revised{different} methods as follows: 
\begin{itemize}
    \item \textbf{Baseline}: A navigation policy trained without \revised{a} latent prediction module, relying solely on the current local height map observations.
    \item \textbf{Zhang~\etal\cite{zhang2024resilient}}: A state-of-the-art local navigation policy that utilizes LSTM to encode memories of past height map observations.
    \item \textbf{Diffusion~\cite{ho2020denoising}}: A navigation policy trained using latent predictions generated \revised{by} a diffusion model.
    \item \textbf{\varTitle \space (Ours)}: Our proposed navigation policy trained using latent predictions generated \revised{by} a \ac{CFM}.
\end{itemize}

% evaluation metrics
We evaluated the navigation policies using three performance metrics: i)~navigation success rate (SR), ii)~collision rate (CR) to measure safe navigation, and iii)~success weighted by path length (SPL)~\cite{anderson2018evaluation} to assess path efficiency. 
A trial is regarded as successful if the robot reaches the goal within $0.4\,\mathrm{m}$ in $20\,\mathrm{s}$.
The CR is defined as the average number of collisions incurred by the non-foot links of the robot.
The SPL is formulated as follows:
\begin{equation}
    \text{SPL} = \frac{1}{N} \sum_{i=1}^{N} S_i \cdot \frac{L_i}{\max(P_i, L_i)},
\end{equation}
where $S_i \in \{0,1\}$, $L_i$, and $P_i$ are the success  rate, \revised{the} optimal path length, and \revised{the} actual path length for \revised{the $i$-th} episode, respectively.

\subsubsection{Real-World Deployment}
Real-world experiments were performed on a Unitree Go2 robot, as shown in \fig{exp_setup}(b), evaluating both the \revised{b}aseline and our proposed method, \varTitle.
We used two Livox Mid-360 LiDARs to obtain point cloud data as the exteroception. 
For the perception module, we employed FAST-LIO~\cite{xu2021fast} for odometry and \revised{Elevation Mapping}~\cite{fankhauser2018probabilistic} for height mapping. The perception and high-level navigation policy were deployed on an Intel NUC Core i7-11700K CPU, while the low-level controller~\cite{nahrendra2023dreamwaq} was deployed on the robot's onboard Jetson Orin NX.

%% file: v4/5_Results_and_Analysis/results.tex
\section{Results and Analysis}

\input{\MainPath/5_Results_and_Analysis/fig_05.tex}
\input{\MainPath/5_Results_and_Analysis/a.latent_prediction}

\input{\MainPath/5_Results_and_Analysis/fig_06.tex}
\input{\MainPath/5_Results_and_Analysis/b.navigation_performance}
\input{\MainPath/5_Results_and_Analysis/c.real_world}

%% file: v4/5_Results_and_Analysis/fig_05.tex
\begin{figure*}[!t]
	\captionsetup{font=footnotesize}
	\centering 
	\includegraphics[width=1.0\textwidth]{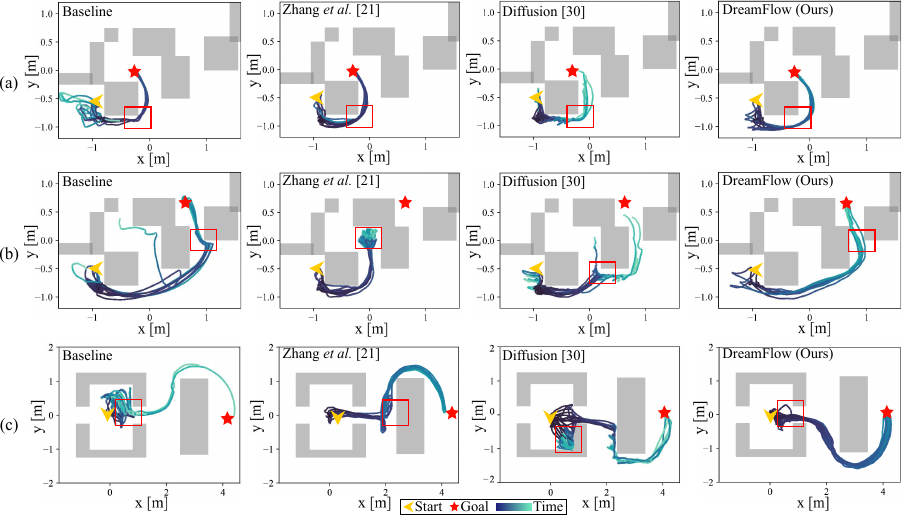}
        \caption{
            Comparison of local navigation performance across three simulation scenarios: (a) Maze (\texttt{Easy}), (b) Maze (\texttt{Hard}), and (c) Hallway.
            Robot trajectories are depicted from a top view and overlaid on a height map, starting from initial positions (yellow markers) and ending at goal locations (red stars).
            Trajectories are color-coded according to time progression, with a gradient from dark blue (start) to cyan (goal).
            \varTitle \space demonstrates smoother trajectories, with better obstacle avoidance and more efficient path selection (highlighted by the red box).
            In contrast, other methods show frequent obstacle contacts and often get stuck in local minima.
        }
        \vspace{-0.3cm} %
	\label{fig:sim}
\end{figure*}

%% file: v4/5_Results_and_Analysis/a.latent_prediction.tex
\begin{table}[!t]
    \centering
    \caption{
    Comparison of latent prediction architectures on \ac{ID} and \ac{OOD} test sets. 
    $\mathrm{Cos.}$ represents cosine similarity between predicted and ground-truth extended latents, where higher values indicate better prediction quality. Inference time is averaged per sample. The best results are indicated in \textbf{bold}.}
    \label{tab:fm_architecture_comparison}
    \renewcommand{\arraystretch}{1.4}
    \setlength{\tabcolsep}{4pt}  
    \small
    \begin{tabular}{l|c|c|c}
    \toprule
    \multirow{2}{*}{Architecture} 
        & \multicolumn{2}{c|}{Cos. $\mathrm{[mean \pm std]}$} 
        & \multirow{2}{*}{Time $\mathrm{[ms]}\,(\downarrow)$} \\
    \cline{2-3}
    & ID & OOD & \\
    \midrule
    MLP                & 0.928 $\pm$ 0.012 & 0.702 $\pm$ 0.136 & \textbf{0.0007} \\
    Diffusion~\cite{ho2020denoising}          & 0.911 $\pm$ 0.349 & 0.919 $\pm$ 0.380 & 0.0218   \\
    FM    & 0.873 $\pm$ 0.020 & 0.784 $\pm$ 0.105 & 0.0117 \\
    CFM (Ours)    & \textbf{0.980} $\pm$ 0.015 & \textbf{0.973} $\pm$ 0.017 & 0.0122 \\
    \bottomrule
    \end{tabular}
    \vspace{-0.2cm}
\end{table}

\subsection{Latent Prediction}\label{sec:results_latent_pred}

Table~\ref{tab:fm_architecture_comparison} summarizes the latent prediction methods on both \ac{ID} and \ac{OOD} test sets. The deterministic MLP performs reasonably well on \ac{ID} data but suffers substantial degradation on \ac{OOD} samples, with sharply increased variance that reveals overfitting and unstable predictions on unfamiliar terrains.
In contrast, flow-based methods exhibit stronger generalization, thanks to \revised{their} generative capability to predict unseen regions.  
Although FM achieves more stable performance on \ac{OOD} compared with MLP, its improvement remains limited, suggesting that flow modeling alone is insufficient without contextual guidance.
Diffusion models~\cite{ho2020denoising}, another class of generative model, demonstrate competitive performance across distributions.
However, their variance remains high and the inference time is considerably longer, as the stochastic iterative sampling procedure accumulates noise across multiple steps.
Consequently, they are less practical for real-time navigation.

The key advantage of \ac{CFM} lies in its conditioning on the robot navigation context, which improves the mean prediction quality and substantially reduces \revised{the} variance, even on \ac{OOD} data. 
By leveraging the navigational context rather than memorizing terrain-specific patterns, \ac{CFM} maintains stable predictions across unseen environments.
% The key advantage of CFM lies in its conditioning on robot state, which improves mean prediction quality while substantially reducing variance, even on OOD data.
% By leveraging robot context rather than memorizing terrain patterns, CFM maintains stable predictions across unseen environments.
% Although MLP provides faster inference through a single forward pass, CFM offers a more favorable balance between accuracy, consistency, and efficiency, remaining well within real-time control constraints.

%% file: v4/5_Results_and_Analysis/fig_06.tex
\begin{figure*}[!t]
	\captionsetup{font=footnotesize}
	\centering 
	\includegraphics[width=1.0\textwidth]{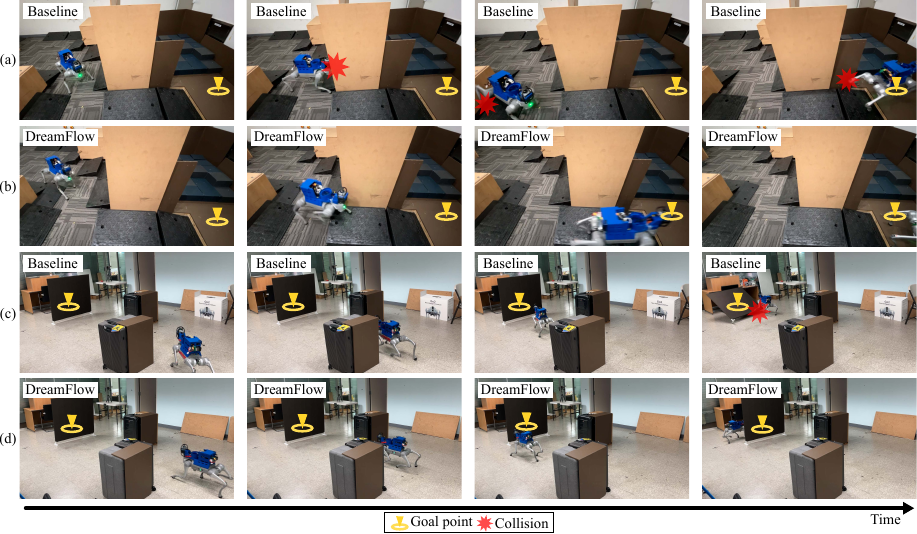}
	\caption{
        Real-world navigation experiments on the quadruped robot, comparing baseline and \varTitle.
        Time progresses from left to right across four snapshots. 
        Yellow and red markers indicate goals and collision events, respectively.
        (a)–(b) Narrow environment with tight passages: the baseline policy frequently collides with walls at corners, whereas \varTitle~achieves collision-free navigation by anticipating corridor layouts beyond immediate perception.
        (c)–(d) Cluttered environment with box obstacles and a wall segment: the baseline collides with the wall due to limited perception, while \varTitle~successfully navigates without collisions through predictive terrain modeling.
        }
        \vspace{-0.3cm} %
	\label{fig:real}
\end{figure*}
\vspace{-5pt}

%% file: v4/5_Results_and_Analysis/b.navigation_performance.tex
% \subsection{Navigation Performance}

% Table~\ref{tab:fm_nav_ablation} presents navigation performance across environments of increasing difficulty. 
% In the \textit{maze} environments, the baseline method achieves reasonable success rates in easier settings but with much higher collision rates than CFM. 
% This indicates that the baseline reaches goals through trajectories with frequent obstacle contacts. 
% CFM maintains high success rates with minimal collisions across both maze difficulties and demonstrates robust navigation through dead ends and trap configurations.

% The \textit{hallway} environment requires spatial reasoning to evaluate whether gaps are traversable given the robot's body dimensions. 
% This environment most clearly differentiates the methods: CFM achieves much higher success rates than all baselines while other methods struggle with the spatial assessment required for narrow passage navigation. 
% The higher SPL of CFM indicates it not only succeeds more often but does so with shorter path lengths relative to the optimal path.

% These results validate that CFM's probabilistic prediction of extended terrain provides the spatial understanding necessary for complex navigation tasks. 
% The performance gap between CFM and baselines widens as environments demand more sophisticated reasoning about unobserved terrain and traversability.

% \vspace{+0.2cm} %
\subsection{Navigation Performance}\label{sec:results_sim}

Table~\ref{tab:fm_nav_ablation} and Fig.~\ref{fig:sim} summarize the navigation performance across Maze and Hallway environments of increasing difficulty. 
In the Maze environments, the \revised{b}aseline policy often relies on redundant exploration, reaching goals only after long detours and frequent obstacle contacts.  
Zhang~\etal~\cite{zhang2024resilient} and \revised{the} Diffusion-based method~\cite{ho2020denoising} suffer from local minima in the \texttt{Hard} scenario, where short-sighted decisions trap the robot and severely reduce success rates.  
In contrast, \varTitle~avoids these local minima by predicting extended terrain, enabling more directed exploration and substantially higher goal-reaching success.  
Furthermore, \varTitle~achieves this with minimal collisions, as the predicted extended latent \revised{representation} allow\revised{s} the policy to anticipate potential contact situations far before the robot \revised{approaches} the obstacles. 

The Hallway environment requires reasoning about traversability of narrow gaps relative to the robot’s body dimensions. 
The compared policies frequently interpret passable corridors as a \revised{single large} obstacle with high risk, hindering the robot \revised{from passing through}. 
However, \varTitle~policy is able to discern this passage and safely navigates the robot through it, resulting in the highest success rate and shortest SPL.

Overall, these results validate that \varTitle~policy with the CFM’s probabilistic prediction of extended terrain equips the policy with foresight that is essential for escaping local minima, reducing redundant exploration, and proactively avoiding collisions in complex environments.

\begin{table}[t!]
    \centering
    \caption{Navigation performance in Maze and Hallway environments. 
    Metrics are success rate (SR), success weighted by path length (SPL), and collision rate (CR). 
    CFM achieves the highest SR and SPL while maintaining the lowest CR, demonstrating its ability to avoid local minima and collisions compared with other methods. The best results are indicated in \textbf{bold}.}
    \label{tab:fm_nav_ablation}
    \renewcommand{\arraystretch}{1.3}
    \setlength{\tabcolsep}{2.5pt} 
    \begin{adjustbox}{max width=\linewidth}
    \begin{tabular}{l|
    S[table-format=2.1]S[table-format=1.2]S[table-format=1.1]|
    S[table-format=2.1]S[table-format=1.2]S[table-format=2.2]|
    S[table-format=2.1]S[table-format=1.2]S[table-format=2.2]
    }
    % \begin{tabular}{l@{\hskip 3pt}|c@{\hskip 0.5pt}cc@{\hskip 0.5pt}|c@{\hskip 0.5pt}cc@{\hskip 0.5pt}|c@{\hskip 0.5pt}cc}
    \toprule
    \multirow{2}{*}{Method}
      & \multicolumn{3}{c|}{Maze (\texttt{Easy})} 
      & \multicolumn{3}{c|}{Maze (\texttt{Hard})}
      & \multicolumn{3}{c}{Hallway} \\
    \cline{2-10}
      & SR$\uparrow$ & SPL$\uparrow$ & CR$\downarrow$
      & SR$\uparrow$ & SPL$\uparrow$ & CR$\downarrow$
      & SR$\uparrow$ & SPL$\uparrow$ & CR$\downarrow$ \\
    \midrule
    Baseline   & 83.2 & 0.23  & 3.9  & 76.5 & 0.37  & 54.8  & 35.8  & 0.21  &  5.1 \\
    % MLP        & 93.9 & 0.31 & 5.8  & 38.3 & 0.18  & 18.2  & 50.6  & 0.37  & 2.5 \\
    Zhang~\etal\cite{zhang2024resilient}       & 95.3 & 0.33 & 2.5  & 5.4  & 0.03  & 15.6  & 25.1  & 0.12  & 4.9 \\
    Diffusion~\cite{ho2020denoising}  & 88.4 & 0.28 & 3.1  & 68.9 & 0.32  & 43.9  & 33.9  & 0.19  & 23.6 \\
    \varTitle~(Ours) & $\mathbf{99.6}$ & $\mathbf{0.35}$ & $\mathbf{0.9}$  
                & $\mathbf{83.1}$ & $\mathbf{0.45}$ & $\mathbf{8.9}$  
                & $\mathbf{89.8}$ & $\mathbf{0.58}$ & $\mathbf{2.3}$ \\
    \bottomrule
    \end{tabular}
    \end{adjustbox}
    \vspace{-0.2cm}
\end{table}

%% file: v4/5_Results_and_Analysis/c.real_world.tex
\subsection{Real World Experiments}\label{sec:results_real}

We validated our approach through real-world experiments in two environments, as shown in Fig.~\ref{fig:real}. 
The \textit{narrow} environment (rows a-b) consists of a maze-like configuration with tight passages that challenge the robot's ability to navigate through confined spaces.
The \textit{cluttered} environment (rows c-d) features two box obstacles and one wall segment that create multiple occlusions between the robot and the target position.

% We compared the baseline method using only local observations against \varTitle. 
As illustrated in Fig.~\ref{fig:real}(a), the baseline method in the narrow environment frequently collides with walls (red markers), particularly when attempting to navigate corners where its limited perception prevents it from identifying traversable paths. 
In contrast, Fig.~\ref{fig:real}(b) shows that \varTitle~achieve\revised{s} collision-free navigation by anticipating corridor layouts beyond its immediate sensor range, smoothly reaching the goal point (yellow marker) through proactive path adjustments.

The cluttered environment further highlights these performance differences. 
As shown in Fig.~\ref{fig:real}(c), the baseline collides with the wall segment that occludes the target position, unable to perceive a viable path around this obstacle until impact. 
Meanwhile, Fig.~\ref{fig:real}(d) demonstrates \varTitle's successful collision-free navigation, where the extended terrain prediction enable\revised{s} the robot to plan an efficient path around the occluding wall from the beginning.

These real-world results confirm that the ability of \varTitle~to predict extended terrain from partial observations translates effectively to physical deployments, where real-world uncertainties such as sensor noise and more severe partial observability exist.
While the baseline's reactive approach led to multiple collisions, our method maintained safe, collision-free trajectories throughout all trials, validating the practical value of predictive terrain modeling for real-world robot navigation.

%% file: v4/6_Conclusion/conclusion.tex
\section{Conclusion}~\label{section:conclusion}
% In this study, we present, DreamFlow, 
% 장점 : CFM은 probabilistic한 특성 때문에 unseen 에 대한 predict에 강점이 있으며, 이는 navigation 성능 향상에 도움이 된다. 
% 실험적으로 redundant Exploration, local minima에 빠지는 횟수, collision이 줄었음을 보였다.  
% 한계점 : 1. 제한적인 실험 환경 (정적) 에서만 평가되어서, moving obstacle에서 어떤 지 평가가 안 됨. 
% 한계점 : 2. training 단계가 좀 많음. 
% Future work : 하나의 Prediction만 사용해서 multiple 한 candidate의 활용 가능성이 있음.
% In this paper, we present\revised{ed} \varTitle, an \ac{DRL}-based framework that integrates \ac{CFM} to enhance local navigation in \revised{a} limited observation range.
% Our experiments demonstrate\revised{d} that incorporating \ac{CFM}-based probabilistic latent prediction, which extrapolates from local observations to infer the extended environmental structure, effectively reduce\revised{d} redundant exploration, alleviate\revised{d} failures from local minima, and decrease\revised{d} collisions in cluttered environments.
% Although these results are promising, there is still room for further research. 
% A more systematic study of prediction horizons could provide deeper insights into the relationship between foresight range and navigation performance.
% % Additionally, extending the evaluation to dynamic environments will be important for validating robustness in more realistic settings.
% For future work, we will integrate the current training pipeline into an end-to-end framework and explore multiple latent predictions to enhance robustness in complex dynamic environments.
In this paper, we present\revised{ed} \varTitle, an \ac{DRL}-based framework that integrates \ac{CFM} to enhance local navigation in \revised{a} limited observation range.
Our experiments demonstrate\revised{d} that \ac{CFM}-based probabilistic latent prediction, which extrapolates from local observations to infer the extended environmental structure, effectively reduce\revised{d} redundant exploration, alleviate\revised{d} failures from local minima, and decrease\revised{d} collisions in cluttered environments.
For future work, we plan to investigate the relationship between prediction horizon and navigation performance, integrate the training pipeline into an end-to-end framework, and explore multiple latent predictions for complex dynamic environments.